\documentclass[10pt,twocolumn,letterpaper]{article}

\usepackage{cvpr}              %
\usepackage[T1]{fontenc}

\usepackage{graphicx}
\usepackage{amsmath}
\usepackage{amssymb}
\usepackage{amsfonts}
\usepackage{booktabs}
\usepackage{xcolor}
\usepackage[multidot]{grffile}

\usepackage{outlines} %

\usepackage[accsupp]{axessibility}  %

\usepackage[pagebackref,breaklinks,colorlinks]{hyperref}

\usepackage[capitalize]{cleveref}
\crefname{section}{Sec.}{Secs.}
\Crefname{section}{Section}{Sections}
\Crefname{table}{Table}{Tables}
\crefname{table}{Tab.}{Tabs.}

\def\cvprPaperID{8832} %
\def\confName{CVPR}
\def\confYear{2022}

\newcommand{\ourhat}[1]{\widehat{#1}}

\def\DSP{HDR-DSP }

\begin{document}

\title{Self-Supervised Super-Resolution for Multi-Exposure Push-Frame Satellites}

\author{Ngoc Long Nguyen$^1$\hspace{0.85 cm} \hfill Jérémy Anger$^{1,2}$\hspace{0.85 cm} \hfill Axel Davy$^1$\hspace{0.85 cm} \hfill  Pablo Arias$^1$\hspace{0.85 cm} \hfill Gabriele Facciolo$^1$\\
$^1$ Université Paris-Saclay, CNRS, ENS Paris-Saclay, Centre Borelli, France \hspace{0.6cm}$^2$ Kayrros SAS\\
{\tt\normalsize \color{purple} \url{https://centreborelli.github.io/HDR-DSP-SR/}}\\
}

\maketitle

\setcounter{page}{1}

\begin{abstract}

Modern Earth observation satellites capture multi-exposure bursts of push-frame images that can be super-resolved via computational means. In this work, we propose a super-resolution method for such multi-exposure sequences, a problem that has received very little attention in the literature. The proposed method can handle the signal-dependent noise in the inputs, process sequences of any length, and be robust to inaccuracies in the exposure times. Furthermore, it can be trained end-to-end with self-supervision, without requiring ground truth high resolution frames, which makes it especially suited to handle real data. Central to our method are three key contributions: i) a base-detail decomposition for handling errors in the exposure times, ii) a noise-level-aware feature encoding for improved fusion of frames with varying signal-to-noise ratio and iii) a permutation invariant fusion strategy by temporal pooling operators. We evaluate the proposed method on synthetic and real data and show that it outperforms by a significant margin existing single-exposure approaches that we adapted to the multi-exposure case.

\end{abstract}

\section{Introduction}
\label{sec:intro}

\begin{figure*}[!htbp]
    \centering
    \def\s{0.095}
    \def\ss{0.195}

    \begin{subfigure}{\s\linewidth}
    \includegraphics[width=\linewidth]{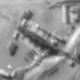}
    \caption*{$LR_1$}
    \end{subfigure}
    \begin{subfigure}{\s\linewidth}
    \includegraphics[width=\linewidth]{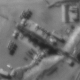}
    \caption*{$LR_2$}
    \end{subfigure}
    \begin{subfigure}{\s\linewidth}
    \includegraphics[width=\linewidth]{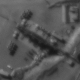}
    \caption*{$LR_3$}
    \end{subfigure}
        \begin{subfigure}{\s\linewidth}
    \includegraphics[width=\linewidth]{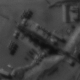}
    \caption*{$LR_4$}
    \end{subfigure}
        \begin{subfigure}{\s\linewidth}
    \includegraphics[width=\linewidth]{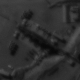}
    \caption*{$LR_5$}
    \end{subfigure}
        \begin{subfigure}{\s\linewidth}
    \includegraphics[width=\linewidth]{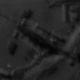}
    \caption*{$LR_6$}
    \end{subfigure}
        \begin{subfigure}{\s\linewidth}
    \includegraphics[width=\linewidth]{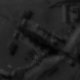}
    \caption*{$LR_7$}
    \end{subfigure}
        \begin{subfigure}{\s\linewidth}
    \includegraphics[width=\linewidth]{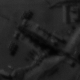}
    \caption*{$LR_8$}
    \end{subfigure}
        \begin{subfigure}{\s\linewidth}
    \includegraphics[width=\linewidth]{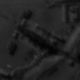}
    \caption*{$LR_9$}
        \end{subfigure}
        \begin{subfigure}{\s\linewidth}
    \includegraphics[width=\linewidth]{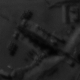}
    \caption*{$LR_{10}$}
    \end{subfigure}
    
    \begin{subfigure}{\ss\linewidth}
    \includegraphics[width=\linewidth]{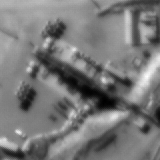}
    \caption*{ME S\&A}
    \end{subfigure}
    \begin{subfigure}{\ss\linewidth}
    \includegraphics[width=\linewidth]{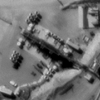}
    \caption*{Planet L1B~\cite{murthy2014SkySat}}
    \end{subfigure}
    \begin{subfigure}{\ss\linewidth}
    \includegraphics[width=\linewidth]{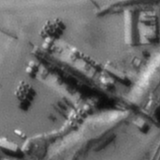}
    \caption*{BD-ACT~\cite{anger2020fast}}
    \end{subfigure}
    \begin{subfigure}{\ss\linewidth}
    \includegraphics[width=\linewidth]{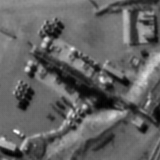}
    \caption*{DSA~\cite{Nguyen_2021_CVPR}}
    \end{subfigure}
    \begin{subfigure}{\ss\linewidth}
    \includegraphics[width=\linewidth]{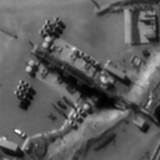}
    \caption*{Our \DSP}
    \end{subfigure}
    \caption{Super-resolution from a real multi-exposure sequence of 10 SkySat images. Top row: Original low resolution images with different exposures. Bottom row: Reconstructions from five methods, including ours trained with self-supervision (right).}\label{fig:new-teaser-real-images}
    \vspace{-1em}
\end{figure*}

High resolution (HR) satellite imagery is a key element in a broad range of tasks, including human activity monitoring and disaster relief.
Super-resolution by computational methods has recently been adopted~\cite{murthy2014SkySat,anger2020fast} by the remote sensing industry (Planet SkySat, Satellogic Aleph-1). 
By leveraging high frame rate low-resolution (LR) acquisitions, low-cost constellations can be effective competitors to more traditional high-cost satellites.
In order to capture the full dynamic range of the scene, some satellites use exposure bracketing, resulting in sequences with varying exposures. While several works have addressed multi-image super-resolution (MISR) of single-exposure sequences, almost no previous work considers the multi-exposure case.

MISR techniques exploit the aliasing in several LR acquisitions to reconstruct a HR image.
The maximum attainable resolution is capped by
the spectral decay of the blur kernel resulting from the sensor's pixel integration and the camera optics, which imposes a frequency cutoff beyond which there is no usable high frequency information.

Aggregating many frames is also interesting as it allows significant noise reduction. If the LR frames are acquired with bracketed exposures, it is possible to integrate them in a super-resolved high dynamic range (HDR) image. Long exposures have higher signal-to-noise ratio (SNR) which helps reduce the noise in dark regions, whereas short exposures provide information in bright regions which can cause saturation with longer exposure times.

In this work, our goal is to perform joint super-resolution and denoising 
from a time series of bracketed satellite images. We focus on push-frame satellite sensors such as the SkySat constellation from Planet. 
We increase the resolution by a factor of two, which is the frequency cutoff of the combined optical and sensor's imaging system.
The SkySat satellites~\cite{murthy2014SkySat} contain a full-frame sensor capable of capturing bursts of overlapping frames: a given point on the ground is seen in several consecutive images. %
However, our technique is general and can be applied to other satellites, or beyond satellite imagery to consumer cameras capable of multi-exposure burst or video acquisition.

Several methods have addressed either MISR or HDR imaging from multiple exposures, but their combination has received little attention.
Existing works consider an ideal setup in which frames can be aligned with an affinity~\cite{traonmilin2014simultaneous, anger2020fast} or a homography~\cite{vasu2018joint}, and the number of acquisitions is large enough to render the problem an overdetermined system of equations. 
Such motion models are good approximations for satellite bursts, but ignore parallax~\cite{anger2021parallax}, which can be noticeable for mountains and tall buildings.

In the case of satellite imaging, push-frame cameras capable of capturing multi-exposure bursts are relatively recent, which explains why all previous works on MISR focus on the single-exposure case~\cite{molini2020deepsum++,deudon2002highres,anger2020fast,Nguyen_2021_CVPR}, except for SkySat's proprietary method~\cite{murthy2014SkySat} producing the L1B product, whose details are not public.
Deep learning methods currently outperform traditional model-based approaches~\cite{rohith2021paradigm}.
In general, learning-based methods require large realistic datasets with ground truth to be trained, as methods trained on synthetic data~\cite{agustsson2017ntire} fail to generalize to real images~\cite{cai2019toward}.
One of such datasets is the PROBA-V dataset~\cite{martens2019super}, acquired with a satellite equipped with two cameras of different resolutions. This dataset has fostered the publication of several deep learning approaches to satellite MISR~\cite{molini2019deepsum, deudon2002highres, arefin2020multi}. %
However, the PROBA-V dataset is not appropriate for MISR of LR image bursts acquired at a high frame rate, as the PROBA-V sequences are multi-date and present significant content and illumination changes.

A promising direction is to use self-supervised learning techniques, which have been applied to video restoration tasks such as denoising and demosaicing~\cite{ehret2019model,ehret2019join,dewil2021self,yu2020joint,sheth2021unsupervised}, and recently to MISR \cite{Nguyen_2021_CVPR}. 
These techniques benefit from the temporal redundancy in videos. Instead of using ground truth labels, one of the degraded frames in the input sequence is withheld from the network and used as label. %

Our work builds upon \emph{Deep Shift-and-Add} (DSA)~\cite{Nguyen_2021_CVPR}, a self-supervised deep learning method for MISR of single-exposure bursts of satellite images. 
The model is trained without supervision by exploiting the frame redundancy.

\smallskip
\noindent\textbf{Contributions.}
In this work, we propose \emph{High Dynamic Range Deep Shift-and-Pool}, \DSP a self-supervised method 
for joint super-resolution and denoising of bracketed satellite imagery.
The method is able to handle time-series with a variable number of frames
and is robust to errors in the exposure times,
as the ones provided
in the
metadata are often inaccurate.
This makes our method directly applicable to real image data (see Figure~\ref{fig:new-teaser-real-images}).
This is, to the best of our knowledge, the first multi-exposure MISR method for satellite imaging, and beyond satellite imagery, it is the first approach based on deep-learning.

Our contributions are the following:

\noindent\emph{Feature Shift-and-Pool.} We propose a \emph{shift-and-pool} module that merges features
(computed by an encoder network on each input LR frame) into a HR feature map by temporal pooling using permutation invariant statistics: average, maximum, and standard deviation. This gives a rich fused representation which yields a substantial improvement over the average~\cite{Nguyen_2021_CVPR}, in both single and multiple exposure cases.

\noindent\emph{Robustness to inaccurate exposure times via base-detail decomposition.}  We propose normalizing the input frames and decomposing them into base and detail.
The errors caused by the inaccurate exposure times affect mainly the base, 
whereas the detail containing the aliasing required for super-resolution can be safely processed by the network. %
Note that vignetting and stray light can also cause exposure issues that affect single and multi-exposure MISR alike.

\noindent\emph{Noise-level-aware detail encodings.} The noise present in the LR images is signal-dependent, its variance being an affine function of the intensity. 
To deal with such noise, we provide the un-normalized LR images to the encoder in addition to the normalized detail components. 
This gives the encoder information about the noise level of each pixel, necessary for an optimal fusion.

\noindent\emph{Self-supervised loss with grid shifting.}
{Using random shifts of the high-resolution grid, we make the self-supervised loss of~\cite{Nguyen_2021_CVPR} translation equivariant, leading to improved results.}

We validate our contributions with an ablation study on a synthetic dataset~(\S\ref{sec:ablation}), designed to {model} the main characteristics of real bracketed SkySat sequences.
Since there are no previous works on multi-exposure MISR, we compare against state-of-the-art single-exposure MISR methods which we adapt and retrain to multi-exposure inputs (\S\ref{sec:comparison-with-sota}). 

We also introduce a dataset of 2500 multi-exposure real SkySat bursts (\S\ref{sec:real-data}). The dataset only consists of noisy LR images, but we can nevertheless train our network on it, since it is self-supervised. 
Both on synthetic and real data, the proposed \DSP method attains the best results by a significant margin \emph{even though it is trained without high resolution ground truth data}.
\underline{The dataset is available for} \underline{download on the \href{https://centreborelli.github.io/HDR-DSP-SR/}{project website}}.

\section{Related work}

Most works on video and burst super-resolution focus on the single-exposure case~\cite{lecouat2021lucas, bhat2021deep,sajjadi2018frame, tao2017detail,anger2020fast,deudon2002highres,molini2019deepsum,Nguyen_2021_CVPR}.
The problem of super-resolution from multi-exposure sequences has received much less attention. 
In~\cite{traonmilin2014simultaneous} it is modeled as an overdetermined system and solved via a non regularized least-squares approach. 
An affine motion model and exact knowledge of the exposure times are assumed. The authors in~\cite{vasu2018joint} address the case in which the images have motion blur due to the camera shake.
They also consider a static scene and do not consider noise.
A related method for HDR imaging uses dual exposure sensors, which interlace two exposures in even and odd columns of the image~\cite{heide2014flexisp,ccougalan2020hdr}. This can be seen as horizontally super-resolving the video.

Other works perform a related task: joint super-resolution and reverse tone-mapping~\cite{kim2018multi,kim2019deep,kim2020jsi}. The difference with our problem is that the input video is a single-exposure LR video, and the goal is to artificially increase its dynamic range to adapt it to HDR screens.

Methods for HDR imaging from multiple exposures need to deal with the noise. Granados et al.~\cite{granados2010optimal} address the case of signal-dependent noise and propose a fixed point iteration of the MLE estimator which is close to the Cramer-Rao bound~\cite{aguerrebere2014best}. In these works, the denoising comes only from the temporal fusion. In~\cite{aguerrebere2013simultaneous,aguerrebere2017bayesian}, this is incorporated in into spatio-temporal patch-based denoisers.

Our work can also be related to burst and video joint denoising and demosaicing~\cite{hasinoff2016burst,wronski2019handheld,ehret2019join}, as demosaicing can be regarded as a super-resolution problem. %

\section{Observation model}

\begin{figure*}
    \centering
    \includegraphics[width=1\linewidth]{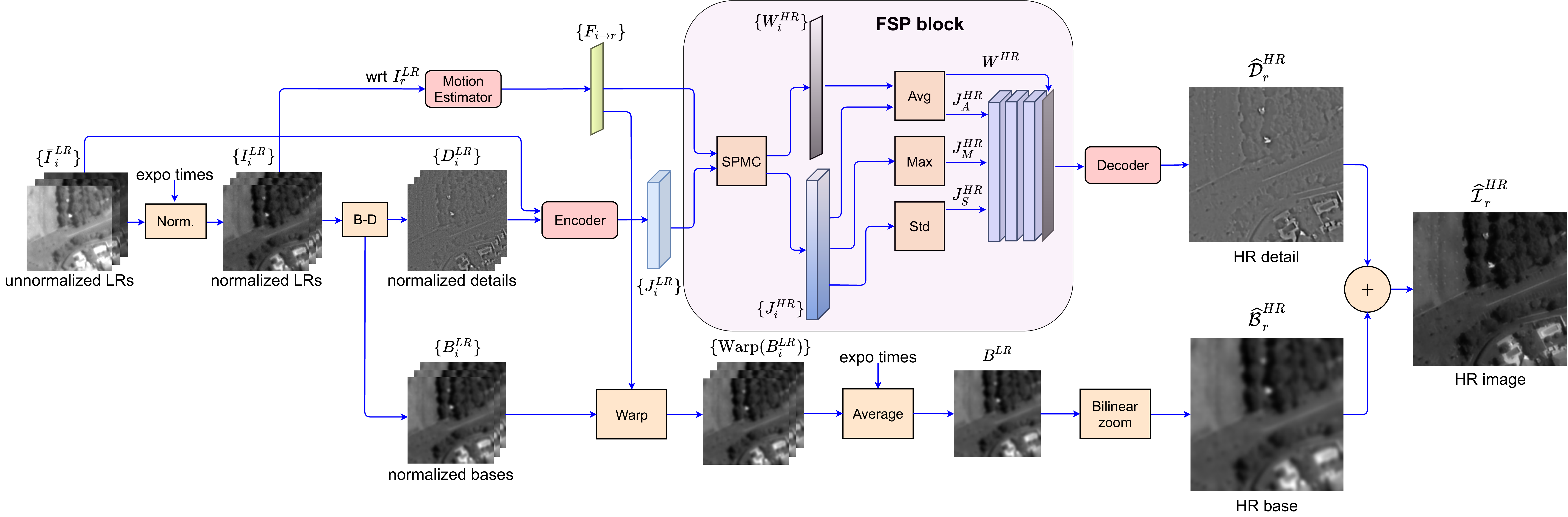}
    \caption{Overview of our proposed multi-exposure super-resolution network architecture \DSP at inference time.}
    \label{fig:Architecture}
\vspace{-1em}
\end{figure*}

We denote by $\mathfrak{I}_t$ a dynamic infinite-resolution ideal scene. The camera on the satellite captures a sequence of $m$ low resolution images $\bar I^{LR}_i$ with different exposures. For the $i-$th acquisition, the dynamic scene $\mathfrak{I}_t$ is integrated during an exposure time $e_i$ centered at $t_i$. 
Even if satellites travel at a very high speed relative to the ground, 
precise electro-optical image stabilization systems (with piezo-electric actuators~\cite{KARGIEMAN2017,Knapp2020} or steering mirrors~\cite{ROBINSON2019})
assure that the observed scene 
$\mathfrak{I}_t$ is mostly constant during the exposure time ($\sim$2ms), which allows us to approximate the temporal integration with a product in our observation model
\begin{equation}
    \bar I^{LR}_i = e_i\Pi_1 \left( \mathfrak{I}_{t_i}\ast k \,\right) + n_i = e_i {\mathcal{I}}^{LR}_i + n_i.
    \label{eq:image_formation}
\end{equation}
Here $k$ is the Point Spread Function (PSF) modeling jointly optical blur and pixel integration, $\Pi_1$ is the bi-dimensional sampling operator due to the sensor array, 
${\mathcal{I}}^{LR}_i$ is the clean low-resolution image corresponding to an exposure of $1$ unit of time
and $n_i$ denotes the noise. Throughout the text, calligraphic fonts $\mathcal{I}_i$ denote  noise-free images and regular fonts $I_i$ noisy ones. 
A bar $\bar{{I_i}} = e_i {{I_i}}$ indicates that the image is multiplied by its exposure time (i.e. as it is acquired by the sensor), while its absence denotes images \emph{normalized} to an exposure time of 1. 
We consider the $r$-th image $\bar{I}_r^{LR}$ in the time series as the \emph{reference}, and without loss of generality we assume its exposure time to be one, $e_r = 1$. 

We model the noise
as spatially independent, additive Gaussian noise with zero mean and signal-dependent variance $n_i(x) \sim \mathcal N(0, \sigma^2(\bar{\mathcal{I}}^{LR}_i(x)))$,
where 
\begin{equation}
\sigma^2(\bar{\mathcal{I}}^{LR}_i(x)) = ae_i\mathcal{I}^{LR}_i(x) + b,
\end{equation}
is an approximation of the Poisson shot noise plus Gaussian readout noise~\cite{ponomarenko2007automatic,foi2008practical},
with parameters $a$ and $b$.

Because of the spectral decay imposed by the pixel integration and optical blur ($k$),
the images $\mathfrak{I}_{t_i}\ast{}k$ are band limited with a cutoff at about twice the sampling rate of the LR images for SkySat.
\emph{Our goal is to increase the resolution by a factor $2$ by estimating $\ourhat{\mathcal{I}}_r^{HR}$, a non-aliased sampling of $\mathfrak{I}_{t_r}\ast{}k$
from several LR observations $\{\bar{I}^{LR}_i\}_{i=1}^m$ with varying exposures
$\{e_i\}_{i=1}^m$.} A sharp super-resolved image can then be recovered by partially deconvolving $k$.

In order for the method to be applicable in practice, it needs to handle time series with a variable number of frames $m$, and to be robust to inaccuracies in the exposure times $e_i$, as the exposure times in the image metadata are only a coarse approximation of the real ones.

\section{Proposed method}

Our method builds upon the DSA method for MISR introduced in~\cite{Nguyen_2021_CVPR}, which can be regarded as a trainable generalization of the traditional shift-and-add (S\&A) algorithms~\cite{fruchter2002drizzle,merino2007super,grycewicz2008focal,alam2000infrared,jia2012method}. A \emph{feature S\&A} is used to fuse feature representations produced from the LR images by an encoder network. A motion estimation network computes the optical flows between each input LR frame and the reference frame. The output of the feature S\&A is a high-resolution aggregated feature map, which is then decoded by another network to produce the output image.

The DSA method could be extended to multi-exposure sequences by applying it to the normalized images $I_i^{LR} = \bar{I}_i^{LR}/e_i$.
This approach however is sub-optimal because it neglects the fact that the normalization alters the noise variance model, and fails if the reported exposure times are inaccurate, which is the case in practice. %

To better exploit multiple exposures, we propose two modifications: (1) A base-detail decomposition, which provides robustness to errors in the exposure times; (2) An encoding of the images that is made dependent on the noise variance, which allows the encoder to weight different contributions according to their signal-to-noise ratio.
In addition, we also propose a new feature pooling fusion intended to capture a richer picture of the encoded features, leading to a substantial improvement in reconstruction quality, both for single and multiple exposure cases.
The resulting network can be trained end-to-end with self-supervision, i.e. without requiring ground truth.

\subsection{Architecture} \label{Sec:Architecture}

Figure~\ref{fig:Architecture} shows a diagram of our proposed architecture
which takes as input a sequence of multi-exposed LR images $\{\bar{I}_i^{LR}\}_{i=1}^m$ along with the corresponding exposure times $e_i$ and produces one super-resolved image  $\ourhat{\mathcal{I}}_r^{HR}$.
The input LR images are first normalized to unit exposure time. The normalized LR images $\{{I}_i^{LR}\}_{i=1}^m$ are then decomposed into base $\{B_i^{LR}\}$ and detail $\{D_i^{LR}\}$ components. The bases contain the low frequencies. We align and average them to reduce the low frequency noise and upsample the result using bilinear zooming to produce the HR base component. %
The LR detail images are fed to a shared convolutional \emph{Encoder} network that outputs a feature representation of each LR image. The features are then merged into a HR feature map by our \emph{shift-and-pool} block (FSP), which aligns the LR features into the HR grid of the reference frame, 
and applies different pooling operations. The pooled features are then concatenated and fed to a \emph{Decoder} CNN module that produces the HR detail image.
The final HR image is obtained by adding the HR base and detail $\ourhat{\mathcal{I}}_r^{HR} = \ourhat{\mathcal{B}}_r^{HR} + \ourhat{\mathcal{D}}_r^{HR}$.

The trainable modules of the proposed architecture (shown in red in Figure~\ref{fig:Architecture}) include the Motion Estimator, the Encoder, and the Decoder.

\smallskip \noindent\textbf{Base-Detail decomposition.}
\label{para:basedetail}
As mentioned above, normalizing a sequence of the frames $\bar{I}_i^{LR}$ by their reported exposures $e_i$ does not result in stable intensity levels across the sequence. This can be due to small errors in $e_i$. However, uncorrected vignetting or stray light also contribute the same effect, %
even in single-exposure imagery. %

\begin{figure}
    \centering
    \def\s{0.49}
    \begin{subfigure}{\s\linewidth}
    \includegraphics[trim={0 130 80 0 }, clip, width=\linewidth]{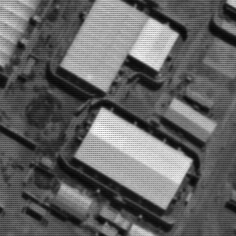}
    \caption*{DSA~\cite{Nguyen_2021_CVPR} (without BD)}
    \end{subfigure}
    \begin{subfigure}{\s\linewidth}
    \includegraphics[trim={0 130 80 0 }, clip, width=\linewidth]{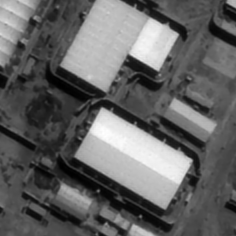}
    \caption*{Our \DSP (with BD)}
    \end{subfigure}
    \caption{High frequency artifacts in a reconstruction  from a real SkySat sequence (using DSA~\cite{Nguyen_2021_CVPR}) with exposure time errors (left). \DSP with the proposed base-detail (BD) decomposition does not present artifacts (right).}
    \label{fig:checkerboard_artifacts_no_basedetail}
\end{figure}

The nature of the super-resolution task makes it very sensitive to these exposure fluctuations. The shift-and-add operation would merge the LR features into an incoherent high-resolution feature map, making the task of the decoder more difficult, resulting in loss of details or high-frequency artifacts (see Figure~\ref{fig:checkerboard_artifacts_no_basedetail}).  
Refining the initial $e_i$ could limit this problem. 
But this entails its own challenges, especially if one also considers vignetting and stray light sources.

Instead, in this paper we propose a more robust and simple alternative, which is based on a base-detail decomposition~\cite{ogden1985pyramid} of the normalized LR images defined as follows
\begin{equation}
\label{eq:B-D}
    B_i^{LR} =  I_i^{LR} \ast G,\quad\quad\quad
    D_i^{LR} =  I_i^{LR} - B_i^{LR},
\end{equation}
for $i = 1,\dots,m$. Here $G$ is a Gaussian kernel of standard deviation 1.
We then process independently the details $\{D_i^{LR}\}$ and the bases $\{B_i^{LR}\}$ to produce the corresponding high resolution estimates $\ourhat{\mathcal{D}}_r^{HR}$ and $\ourhat{\mathcal{B}}_r^{HR}$. This decomposition is linear and does not affect the super-resolution since the alias is preserved in the detail components $\{D_i^{LR}\}$.

As the detail images span a smaller intensity range than the complete image $I_i^{LR}$, an error $\delta$ in the exposure time results in a small deviation in the detail and a large one in the base: 
$\delta \, B_i^{LR} + \delta \, D_i^{LR} = \delta \, I_i^{LR}.$ 
The small error in the detail can be handled by {a super-resolution method}. %

On the other hand, the base images do not need to be super-resolved, but still need to be denoised. In this work we propose a simple processing that aligns and averages the bases and upsamples the result. To fully exploit the high signal-to-noise ratio of longer exposures, the average is weighted by the exposure times $e_i$
\begin{equation}
B^{HR} = \operatorname{Zoom} \left( \frac{\sum_i e_i \operatorname{Warp}(B_i^{LR})}{\sum_i e_i} \right).
\end{equation} 
This weighting is an approximation of the ML estimator of Granados et al.~\cite{granados2010optimal} (details in the supplementary material).

Base and detail decompositions have been used in super-resolution networks~\cite{kim2019deep,Isobe2020} to focus the network capacity on the details. In our case, the decomposition also provides robustness to errors in the radiometric normalization.

\smallskip \noindent\textbf{Motion Estimator.} 
We follow the works of~\cite{sajjadi2018frame, Nguyen_2021_CVPR}  to build a network (with the same hourglass architecture) that estimates the optical flows between the normalized LR frames $\{I_i^{LR}\}_{i=1}^m$ and the normalized reference frame $I_r^{LR}$
\begin{equation}
    F_{i \to r}\!=\!\textbf{MotionEst}(I_i^{LR}, I_r^{LR}; \Theta_{\textbf{M}})\!\in\![-R,R]^{H\!\times\!W\!\times\!2},
\end{equation} 
{where $\Theta_{\textbf{M}}$ denotes the network parameters.} %
A small Gaussian filter ($\sigma = 1$) is applied to the input images to reduce the alias~\cite{Vandewalle2007, Nguyen_2021_CVPR}.
The network is trained with a maximum motion range of $[-R,R]^2$ (with $R = 5$ pixels). The training was adapted to better handle the noise difference due to the multi-exposure setting (see \S\ref{Sec:Self-SR}).

\smallskip \noindent\textbf{Noise-level-aware detail encodings.}
{
The Encoder module generates relevant features $J_i^{LR}$ for each  normalized LR detail image $D_i^{LR}$ in the sequence
\begin{equation}
    J_i^{LR} = \textbf{Encoder}(D_i^{LR}, \bar{I}_i^{LR}; \Theta_{\textbf{E}}) \in \mathbb{R}^{H\times W \times N},
\end{equation}
where $\Theta_{\textbf{E}}$ is the set of parameters of the encoder and $N=64$ is the number of produced features. 
The network architecture is detailed in the supplementary material.

The un-normalized low resolution frames $\bar{I}^{LR}_i$ are also fed to the encoder. This is motivated by the fact that the maximum likelihood fusion of noisy 
acquisitions into a (HDR) image is a weighted average, where the weights 
are the inverse of the noise variances \cite{granados2010optimal,aguerrebere2014best}}.
In the proposed architecture, the normalized details $D^{LR}_i$ are fused to produce a high resolution detail $\ourhat{\mathcal{D}}_r^{HR}$. 
The noisy un-normalized images are unbiased estimators of an affine function of the noise variances $\sigma^2(\tilde I_i^{LR})/a - b/a$, thus they provide to the encoder the information required to compute the optimal fusion weights. The resulting features $J_i^{LR}$ are then aggregated via a set of pooling operations, without any particular handling related to different source exposures. 

\smallskip \noindent\textbf{Feature Pooling.}
We propose the Feature Shift-and-Pool block (FSP) which maps the LR features into their positions on the reference HR grid and pools them. First the features are ``splatted'' bilinearly onto
the HR grid by the SPMC module~\cite{tao2017detail}. Each LR frame is upscaled by introducing zeros between samples and motion compensated following the flows $F_{i\to r}$. This is differentiable with respect to the intensities and the optical flows.
Each splatted pixel is assigned a bilinear weight depending on the fractional part of
its position
in the HR grid. See \cite{tao2017detail,Nguyen_2021_CVPR} for details.

This results in a set of aligned sparse HR feature maps
\begin{equation}
    J_i^{HR} = \text{SPMC} ( J_i^{LR}, \{F_{i \to r}\} ) \in \mathbb{R}^{sH\times sW \times N},
\end{equation}
and the corresponding bilinear splatting weights $W_i^{HR} = \text{SPMC} ( 1, \{F_{i \to r}\} )$.
The upscaling factor $s$ is set to 2.

As in~\cite{Nguyen_2021_CVPR}, we 
use a weighted average pooling in the temporal direction \eqref{eq:averaging}.
In addition, we propose computing the standard deviation and the max \eqref{eq:stdmax}:
\begin{align}
\textstyle    J_A^{HR} &= 
    ({\sum_i J_i^{HR}})({\sum_i W_i^{HR}})^{-1}, \label{eq:averaging}\\
\textstyle    J_M^{HR} &= 
    {\underset{i}{\operatorname{max}\,} J_i^{HR}},\quad\quad
\textstyle    J_S^{HR} = \underset{i}{\operatorname{std}\,} J_i^{HR}.
\label{eq:stdmax}
\end{align}
Note that this block does not have any trainable parameters, a trainable layer may attain a similar performance at a much higher computational cost (see the supplementary material).

These feature pooling operations render the architecture invariant to permutations of the input frames~\cite{aittala2018burst}.
The key idea is that through end-to-end training, the encoder network will learn to output features for which the pooling is meaningful. 
{Therefore, it is essential that the pooling operation is capable of passing all the necessary information to the decoder.}
Indeed, average pooling captures a consensus of the features, which amounts to a temporal denoising. But in aliased image sequences, it is common to come across features that are only visible in a single frame. Thus, the idea of the max-pooling operation is to preserve these unique features that would otherwise be lost in the average. The standard deviation pooling completes the picture  by measuring the point-wise variability of the features. %

The pooled features are independent of the number of processed frames. But this information is important as the decoder may interpret features resulting from aggregating many images differently than those resulting from
just a few.
For this reason, the aggregation weights $W^{HR} = \sum_i W_i^{HR}$ are also concatenated with the pooled features.
As we will see in \S\ref{sec:ablation},
incorporating $W^{HR}$ improves the network ability to handle a variable number of input frames.

\smallskip \noindent\textbf{Decoder.} The Decoder network reconstructs the HR detail image $\ourhat{\mathcal{D}}_r^{HR}$ from the pooled features
\begin{equation}
    \ourhat{\mathcal{D}}_r^{HR}\!=\!\textbf{Decoder}(J_A^{HR},J_M^{HR},J_S^{HR},W^{HR}; \Theta_{\textbf{D}})\!\in\!\mathbb{R}^{sH\times sW},
\end{equation}
where $\Theta_{\textbf{D}}$ denotes the set of parameters of the decoder. 
The architecture is detailed in the supplementary material.

\subsection{Self-supervised learning} \label{Sec:Self-SR}

To train the \DSP detail fusion network, we adapt the fully self-supervised framework of~\cite{Nguyen_2021_CVPR}, which requires no ground truth HR images.
During training, the LR frames are randomly selected and for every sequence, one frame is set apart as the reference $I^{LR}_r$. Then, all the other LR images in each sequence are registered against the reference using the $\textbf{MotionEst}$ network yielding the flows $F_{i\to r}$. The reference frame serves as the target for the self-supervised training similarly to noise-to-noise~\cite{lehtinen2018noise2noise,ehret2019model}. 
The procedure relies on the minimization of a reconstruction loss in the LR domain plus a motion estimation loss to ensure accurate alignment of the frames. The losses and the proposed adaptations are detailed in the following paragraphs.

\smallskip
\noindent\textbf{Self-supervised SR loss.} The self-supervised loss forces the network to produce an HR detail $\ourhat{\mathcal{D}}_r^{HR}$ 
such that when subsampled, it coincides (modulo the noise) with the withheld target detail $D_r^{LR}$ 
\begin{multline}
    \ell_{self}(\ourhat{\mathcal{D}}_r^{HR}, D_r^{LR}) = \|\Pi_{2}(\ourhat{\mathcal{D}}_r^{HR} \ast k )- D_r^{LR}\|_1,
    \label{eq:ssloss}
\end{multline}
where $\ourhat{\mathcal{D}}_r^{HR} = \textbf{Net}(\{D_i^{LR}\}_{i\neq r}, \{\bar{I}_i^{LR}\}_{i=1}^m)$ is the SR output, and $\Pi_{2}$ is the subsampling operator that takes one pixel over two in each direction. %
As in~\cite{Nguyen_2021_CVPR} we %
{include the convolution kernel $k$ in the loss}.
This forces the network to produce a %
{deconvolved} HR image that once convolved with $k$ and subsampled matches the optical blur present in $D_r^{LR}$.

During training, the LR reference is only used in the motion estimator to compute the optical flows, but it is not fused into the HR result to avoid unwanted trivial solutions~\cite{Batson2019,dewil2021self,Nguyen_2021_CVPR}.
At inference time we use the reference as this leads to improved results \cite{Nguyen_2021_CVPR}.

\smallskip
\noindent\textbf{Grid shifting.} \label{para:jitter}
The self-supervised loss~\eqref{eq:ssloss} downsamples the super-resolved {detail} to compare it with the reference LR {detail}. But since the downsampling is fixed, only the sampled positions intervene in the loss, which breaks the translation equivariance of the method.
To avoid this issue, during training we augment the data by adding to the estimated optical flows a random shift of $0.5\epsilon$ in each dimension ($\epsilon \in \{0,1\}$). As a result, the super-resolved image is shifted by $\epsilon$, which is easily compensated before computing the loss. This yields an improvement in PSNR of 0.2dB.

\smallskip 
\noindent\textbf{Motion estimation loss.}  \label{para:meloss}
The motion estimator is trained with unsupervised learning as in~\cite{jason2016back}. The loss consists of a warping term and a regularization term. 
We observed that the optical flow is very sensitive to the intensity fluctuations between frames
(as in our normalized LR frames $I_i^{LR}$),
which result in imprecise alignments.
To prevent this issue we compute the warping loss on the details rather than on the images, which is common in traditional optical flow~\cite{sajjadi2018frame, Liu_2019_CVPR} . 
The loss is computed for each flow $F_{i\to r}$ %
estimated by the $\textbf{MotionEst}$ module
\begin{multline}
\textstyle      \ell_{me}(\{F_{i\to r}\}_{i=1}^m) =  \lambda_1 TV(F_{i\to r}) + \\
\sum_i \| \textbf{Detail}\left(I_i^{LR} - \textbf{Pullback} ( I_r^{LR}, F_{i\to r} ) \right) \|_1,
\label{eq:meloss}
\end{multline}
where $\textbf{Pullback}$ computes a  bicubic warping of $I_r^{LR}$ according to a flow, $\textbf{Detail}$ applies a high-pass filter, TV is the finite difference discretization of the classic Total Variation regularizer~\cite{rudin1992nonlinear}, and $\lambda_1=0.003$ is a hyperparameter controlling the regularization strength.

\smallskip \noindent\textbf{Training.} The self-supervised training of \DSP is done in two stages. We first pretrain the motion estimator on the simulated data to ensure that it produces accurate flows. Then, we train the entire system end-to-end with the pretrained $\textbf{MotionEst}$ using the self-supervised loss ($\lambda_2=3$)
\begin{equation}
    \text{loss} = \ell_{self} + \lambda_2 \ell_{me}.\label{eq:total-loss-self}
\end{equation}
Other training details are in the supplementary material.

\section{Experiments}\label{sec:experiments}

For our experiments, we use real multi-exposure push-frame images (L1A) acquired by SkySat satellites~\cite{murthy2014SkySat}. For the quantitative evaluations we also simulated a multi-exposure and a single-exposure datasets from
L1B products (super-resolved products by Planet with a factor of 1.25).

\subsection{Simulated multi-exposure dataset} \label{sec:simulated}

The two simulated datasets were generated from 1371 crops of L1B products (1096 train, 200 test,  75 val). First, we generate the noise-free LR images normalized to an exposure time of 1. Random subpixel translations of $\{\Delta_i\}_{i=1}^m$ are applied to the ground truth followed by $\times 2$ subsampling
\begin{equation}
\label{eq:simulated}
\begin{aligned}
    \mathcal I_r^{LR} &=  \Pi_{2} (\mathcal I^{HR}), \\
    \mathcal I_i^{LR} &=  \Pi_{2} (\text{Shift}_{\Delta_i} (\mathcal I^{HR}) ), \quad \quad i \neq r
\end{aligned}
\end{equation}
where $\Pi_{2}$ is the subsampling operator. 
The exposure times are simulated as %
$e_i = \alpha^{c_i}$, where $c_i \in\{-5,..,5\}$, and $\alpha =  \text{uniform}(1.2, 1.4)$.
The noises $n_i = \sqrt{ae_i \mathcal{I}_i^{LR} + b} \, \mathcal N(0,1)$ are then added to all the un-normalized frames to produce the noisy multi-exposure sequence $\bar{I}^{LR}_i = e_i \mathcal{I}_i^{LR} + n_i$. 
The  constants $a=0.119
, b = 12.050$ were estimated from real SkySat images with the Ponomarenko noise curve estimation method~\cite{colom2013analysis, ponomarenko2007automatic}. The single-exposure dataset is generated in the same manner but with all $e_i = 1$.
To simulate the exposure inaccuracies, during training and testing the $e_i$ values are contaminated with noise within a range of 5\%.%

We use a PSNR score in our evaluation. The SkySat L1A images have a dynamic range of 12 bits, but we observed that the peak signal is at about 3400 DN. Therefore, our PSNR is normalized with a peak of 3400. We denote PSNR ME (resp. PSNR SE) as the average PSNRs computed on all the multi-exposure (resp. single-exposure) test sequences.

\subsection{Ablation study} \label{sec:ablation}

We study in Table~\ref{tab:ablation-multiexp} the importance of the base-detail decomposition. We consider simulated multi-exposure (ME) and single-exposure (SE) sequences presenting small exposure errors that match the ones observed in real sequences. If we train \DSP without the proposed base-detail (w/o BD), the performance drops noticeably, which is also visible on real sequences (Figure~\ref{fig:checkerboard_artifacts_no_basedetail}). Even when training specifically for a single-exposure setting, as with DSA~\cite{Nguyen_2021_CVPR}, the performance with base-detail is superior. In addition, we can see that removing the un-normalized LR frame
from the encoder
inputs (w/o LR)
leads to a large performance drop
for both single- and multi-exposure.

\begin{table}%
\begin{center}
\caption{Handling of multi-exposure sequences with base-detail decomposition (BD)  and using the un-normalized LR frames $I_i^{LR}$ as an additional encoder input. }\label{tab:ablation-multiexp}
\setlength{\tabcolsep}{6pt}
\resizebox{\columnwidth}{!}{%
\begin{tabular}{l cccc}
\toprule
Methods (all \DSP) & full &  w/o BD &  w/o BD (trained SE) &   w/o LR  \\
\midrule
PSNR(dB) ME           & \textbf{54.70} & 53.76 & 52.91 & 53.94    \\ 
PSNR(dB) SE           & \textbf{54.72} & 54.16 & 54.54 & 54.16  \\ 
\bottomrule
\end{tabular}
}
\end{center}
\end{table}

The experiment shown in Table~\ref{tab:feature-pooling} studies the impact of using multiple feature pooling strategies: average, maximum, and standard deviation. It shows that using the three  greatly improves the results: about $0.5$dB with respect to just using average. We observed that not including the average among the pooling strategies yields much worse results.

\begin{table}%
\begin{center}

\caption{Feature pooling choice. Using average~(A), maximum~(M), and standard deviation~(S) pooling improves the results.}\label{tab:feature-pooling}

\setlength{\tabcolsep}{6pt}
\resizebox{0.8\columnwidth}{!}{%

\begin{tabular}{l cccc}
\toprule
Features & AMS (\DSP) & AS & AM & A \\
\midrule
PSNR(dB) ME & \textbf{54.70} & 54.46 & 54.44 & 54.17 \\ 
PSNR(dB) SE & \textbf{54.72} & 54.47 & 54.48 & 54.20 \\ 
\bottomrule
\end{tabular}
}
\end{center}
\vspace{-1em}
\end{table}

The aggregation weight feature $W^{HR}$ was added  to improve the handling by the decoder of sequences with variable number of input frames. The results in Table~\ref{tab:ablation-variable-frames} confirm the importance of providing these weights. We also compare with networks trained for a fixed number of frames (\DSP 4 and 14)  and observe that in this case the performance drops even when testing for those specific configurations. We conclude that the weights become useless if the training does not consider a variable number of frames.

\begin{table}%
\begin{center}
\caption{Handling variable number of frames (PSNR ME (dB)).}\label{tab:ablation-variable-frames}

\resizebox{\columnwidth}{!}{%
\begin{tabular}{l cccc}
\toprule
Methods (all \DSP) & full &  w/o $W^{HR}$ & \DSP 4 & \DSP 14\\
\midrule
4 frames & \textbf{52.81} & 52.60 & 52.69 & 51.31 \\ 
14 frames & \textbf{55.85} & 55.59 & 54.26 & 55.53 \\ 
variable $n$ frames & \textbf{54.70} & 54.45 & 53.85 & 54.07 \\ 

\bottomrule
\end{tabular}
}
\end{center}
\vspace{-1em}
\end{table}

Lastly, removing the grid shifting (\S\ref{para:jitter}) from the training also reduces the PSNR ME: from 54.70 to 54.49dB.

\begin{table}%
\begin{center}
\caption{PSNR ME (dB) over the synthetic test set with 15 images in the case of 0\%, 5\% and 20\% exposure time errors.}
\label{tab:comparison}
\resizebox{\columnwidth}{!}{%
\begin{tabular}{l cccccc}
\toprule
Methods & RAMS & ME S\&A & HR-net & BD-ACT & DSA &\DSP\\
\midrule
0\%  exp. error & 52.05 & 53.33 & 54.30 & 54.24 &55.55 & \textbf{56.00} \\ 
5\%  exp. error & 51.84 & 52.43 & 54.22 & 54.23 &54.99 & \textbf{55.99} \\ 
20\% exp. error & 49.95 & 49.19 & 53.82 & 54.20 & 54.30 & \textbf{55.90}\\

\bottomrule
\end{tabular}
}
\end{center}
\vspace{-1em}
\end{table}

\begin{figure*}
    \centering
    \def\s{0.12}
    \begin{subfigure}{\s\linewidth}
    \includegraphics[width=\linewidth]{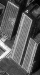}
    \caption*{LR}
    \end{subfigure}
    \begin{subfigure}{\s\linewidth}
    \includegraphics[width=\linewidth]{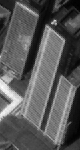}
    \caption*{ME S\&A 43.16dB}
    \end{subfigure}
    \begin{subfigure}{\s\linewidth}
    \includegraphics[width=\linewidth]{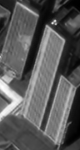}
    \caption*{RAMS 42.99dB}
    \end{subfigure}
    \begin{subfigure}{\s\linewidth}
    \includegraphics[width=\linewidth]{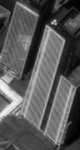}
    \caption*{HR-net 43.58dB}
    \end{subfigure}
    \begin{subfigure}{\s\linewidth}
    \includegraphics[width=\linewidth]{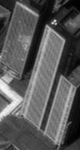}
    \caption*{BD-ACT 43.59dB}
    \end{subfigure}
        \begin{subfigure}{\s\linewidth}
    \includegraphics[width=\linewidth]{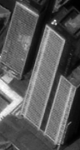}
    \caption*{DSA 45.95dB}
    \end{subfigure}
        \begin{subfigure}{\s\linewidth}
    \includegraphics[width=\linewidth]{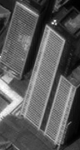}
    \caption*{\DSP \bf 49.32dB}
    \end{subfigure}
        \begin{subfigure}{\s\linewidth}
    \includegraphics[width=\linewidth]{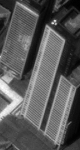}
    \caption*{HR}
    \end{subfigure}
    \caption{Super-resolution from a synthetic multi-exposure sequence (5\% exp.  error) of 15 aliased LR images. Methods are trained on a synthetic dataset and receive as inputs the normalized ME images except BD-ACT and HDR-DSP, which use the base-detail decomposition.}
    \label{fig:comparison_Synthetic}
    \vspace{-1em}
\end{figure*}

\begin{figure}
\captionsetup[subfigure]{labelformat=empty}
    \centering
    \def\s{0.24}%

    \begin{subfigure}{\s\linewidth}%
        \includegraphics[width = \linewidth, trim={0 4 0 8}, clip]{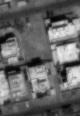}%
        \caption{LR $e = 1.5$}
    \end{subfigure}
    \begin{subfigure}{\s\linewidth}%
        \includegraphics[width = \linewidth, trim={0 4 0 8}, clip]{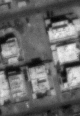}%
        \caption{LR $e = 1.1$}
    \end{subfigure}
    \begin{subfigure}{\s\linewidth}%
        \includegraphics[width = \linewidth, trim={0 4 0 8}, clip]{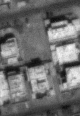}%
        \caption{LR $e = 0.7$}
    \end{subfigure}
    \begin{subfigure}{\s\linewidth}%
        \includegraphics[width = \linewidth, trim={0 4 0 8}, clip]{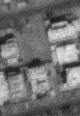}%
        \caption{LR $e = 0.5$}
    \end{subfigure}
    
    \begin{subfigure}{\s\linewidth}%
        \includegraphics[width= \linewidth, trim={0 5 0 10}, clip]{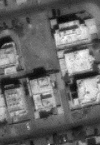}%
        \caption{Planet L1B}
    \end{subfigure}
    \begin{subfigure}{\s\linewidth}%
        \includegraphics[width = \linewidth, trim={0 8 0 16}, clip]{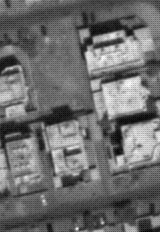}%
        \caption{DSA}
    \end{subfigure}
    \begin{subfigure}{\s\linewidth}%
        \includegraphics[width = \linewidth, trim={0 8 0 16}, clip]{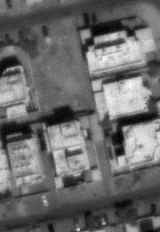}%
        \caption{BD-ACT}
    \end{subfigure}
    \begin{subfigure}{\s\linewidth}%
        \includegraphics[width = \linewidth, trim={0 8 0 16}, clip]{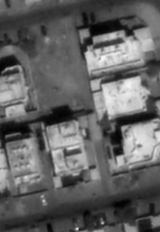}%
        \caption{Our \DSP}
    \end{subfigure}

    \caption{Super-resolution from a real multi-exposure sequence of 9 SkySat images. The first line corresponds to 4 normalized LR images in that sequence with different exposure times. The second line shows the reconstructions by Planet (L1B), DSA, BD-ACT and our method \DSP.}
    \label{fig:comparison_Real}
        \vspace{-1em}
\end{figure}

\subsection{Comparison with the state-of-the-art}
\label{sec:comparison-with-sota}

We compare our self-supervised network on the simulated dataset against state-of-the-art MISR methods for satellite images: \textit{DSA}~\cite{Nguyen_2021_CVPR}, \textit{HighRes-net} (HR-net)~\cite{deudon2002highres}, \textit{RAMS}~\cite{salvetti2020multi}, and \textit{ACT}~\cite{anger2020fast}. A weighted \textit{Shift-and-add}~\cite{merino2007super}  with bicubic splatting adapted to multi-exposure sequences (ME S\&A) serves as the baseline. HR-net and RAMS are two supervised networks designed to perform super-resolution of multi-temporal PROBA-V satellite images. In the context of push-frame satellites, we use the reference-aware version~\cite{nguyen2021proba} of HR-net and RAMS rather than the original approaches, as they achieve higher quality results.  DSA and ACT are two state-of-the-art super-resolution methods for SkySat imagery. ACT also serves as a proxy for comparison with other interpolation-based methods from the literature~\cite{wronski2019handheld}.

We adapt these methods to multi-exposure sequences. The deep learning approaches are fed with the normalized input images, whereas for ACT method we apply the same base-detail decomposition described in \S\ref{para:basedetail} and use ACT to restore the details (denoted BD-ACT). The registration step of ME S\&A, BD-ACT, and RAMS are done with the inverse compositional algorithm~\cite{baker2001equivalence, briand2018improvements}, which is robust to noise and brightness changes. The motion estimator of DSA is also trained with the loss on the details (\S\ref{para:meloss}).

Table~\ref{tab:comparison} shows a quantitative comparison of the methods over the test set in the case of adding exposure time errors of 5\%  (as during training) and 20\%. These errors are estimated from %
SkySat data (exposures ranging from 0.5 to 4.5 ms); see the supplementary material for details. Note that even with exact exposure times (row 0\%), vignetting or stray light effects still justify the use of the proposed base-detail decomposition. Our self-supervised network ranks first in all cases with a significant gain of more than 1dB over all others (see Figure~\ref{fig:comparison_Synthetic}). Interestingly, the performance of most methods degrades quickly for large inaccuracy in exposure times. Only the methods using the base-detail decomposition (BD-ACT and ours) are robust to these inaccuracies. Note that \DSP has never seen errors of 20\%  during training.

\subsection{Results on real data}
\label{sec:real-data}

The proposed self-supervised training allows to train \DSP on real multi-exposure sequences taken from SkySat satellites.
From the L1A product of Planet SkySat, we extracted
2500 sequences ($128 \times 128$ pixels) pre-registered up-to an integer translation.
Out of 2500 sequences, 300 are used for testing.
Each sequence contains from 4 to 15 frames. In about 75\% of the sequences the exposure time varies within each sequence and we used the exposure time information provided in the metadata.

Figure~\ref{fig:comparison_Real} compares \DSP against Planet L1B, DSA, and BD-ACT. The top row shows four normalized frames of the sequence, where we can notice the dependence of the noise level on the exposure time. %
The method used in the Planet L1B product is unknown. It super-resolves by a factor of 1.25 but contains noticeable artifacts and lacks fine details.
The result from DSA exhibits a high-frequency pattern due to the imprecise exposure times.
BD-ACT is able to cope with the exposure changes thanks to the base-detail decomposition, but the result is still very noisy.
In contrast, \DSP shows a clean and detailed reconstruction.

Figure~\ref{fig:new-teaser-real-images} also shows a multi-exposure LR sequence along with the results from ME S\&A, Planet L1B, ACT, DSA and \DSP\!\!.
Comparing \DSP with DSA, we see that the former provides a cleaner result thanks to the base-detail decomposition and the proposed improvements over the DSA architecture and training procedure, which is also observed in the synthetic experiments.

\section{Conclusion and limitations}

The proposed \DSP method is able to reconstruct high-quality results from multi-exposure bursts, providing fine details, low-noise, and high dynamic range. The proposed base-detail processing allows robustness to errors in the exposure time that are common in practice. In addition, a significant performance improvement is obtained by making the image encoding dependent on the noise variance, and using a new feature pooling designed to capture richer representations.
Thanks to its fully self-supervised training, the method requires no ground truth and can thus be applied on real data.
We show its effectiveness by training a model that super-resolves multi-exposure SkySat L1A acquisitions, leading to a substantial resolution gain with respect to the state-of-the-art.

\smallskip \noindent\textbf{Limitations.} 
The context of remote sensing allows one to make additional assumptions that do not hold in more general settings: 
1. The considered noise levels are away from the challenging photon-limited regime;  
2. Motion and occlusions are much easier to handle. 
In particular, the latter point should be improved to apply this method to video or burst super-resolution.
Besides, the proposed method does not handle saturation.
This will be studied in future work.

\paragraph{Acknowledgments.}
Work supported by a grant from Région Île-de-France. %
This work was performed using HPC resources 
from GENCI–IDRIS (grants 2022-AD011012453R1 and 2022-AD011012458R1) and  from the “Mésocentre” computing center of CentraleSupélec and ENS %
Paris-Saclay supported by CNRS and Région Île-de-France (http://mesocentre.centralesupelec.fr/). 
We thank Planet for providing the L1A SkySat images.

{\small
\bibliographystyle{ieee_fullname}
\bibliography{egbib}
}

\end{document}


\maketitle

\section{Weights for the base fusion}
\label{Sec:granados}

Since the base component only contains low frequencies and cannot be super-resolved, we propose a simple pipeline consisting of \emph{i)} alignment of the LR base components $B_i$ to the reference, \emph{ii)} temporal fusion via weighted average to attenuate noise, \emph{iii)} upscaling using bilinear interpolation. For the temporal fusion the weights in the weighted average are simply the exposure times:
\begin{equation}
B^{LR}(x) = \frac{\sum_i e_i \text{Warp}(B^{LR}_i(x))}{\sum_i e_i}
\end{equation}
In this section we will provide a justification for this choice,
which is based on two approximations.

\paragraph{Approximate noise model for the base.} The base results from the convolution with a Gaussian kernel $G$. At pixel $x$ we have
\[B_i^{LR}(x) = \sum_{h} G(h)I_i^{LR}(x+h).\]
Assuming the signal-dependent Gaussian noise model of Eq. (2)\footnote{Tables,
figures and equations in the supplementary material are labeled S1, S2, \dots
to differentiate them from references to the main paper.}, we have
that $B_i^{LR}(x)$ also follows a Gaussian distribution with the following mean and variance:
\begin{align*}
\mathbb E\{B_i^{LR}(x)\} &= \sum_{h} G(h)\mathcal I_i^{LR}(x+h)\\
\mathbb V\{B_i^{LR}(x)\} &= \frac{a}{e_i}\sum_h G^2(h)\mathcal I_i^{LR}(x+h) + \frac{b}{e_i^2}\sum_h G^2(h).
\end{align*}
We are going to assume that the clean LR image $\mathcal I_i^{LR}$ varies smoothly in the filter support, and thus 
\begin{equation}
\mathbb E\{B_i^{LR}(x)\} \approx \mathcal I_i^{LR}(x),\quad
\mathbb V\{B_i^{LR}(x)\} \approx \frac{\alpha e_i\mathcal I_i^{LR}(x) + \beta}{e_i^2}.
\end{equation}
where $\alpha = a\sum_h G^2(h)$ and $\beta = b\sum_h G^2(h)$.
This rough approximation allows us to use a signal-dependent Gaussian noise model like (2). The approximation is only valid in regions where the image is smooth (away from edges, textures, etc.). However, these are the regions in which we are mainly interested, since it is where the low frequency noise present in the base becomes more noticeable.

\paragraph{Approximate MLE estimator for the weights.} After alignment, for a
given pixel $x$ we have different values acquired with varying exposure times,
which we are going to denote as $z_i = \text{Warp}(B_i^{LR}(x))$ to simplify
notation. We also have the corresponding clean $LR$ base images $\mathcal
B^{LR}_i$, and we are going to assume that they coincide after alignment, i.e.
$y = \text{Warp}(\mathcal B_i^{LR})(x)$ for $i = 1,...,m$. We would like to
estimate $y$ from the series of observations
\[z_i \sim\mathcal N\left(y, \sigma_i^2(y)\right), \quad \sigma_i^2(y) = \frac{\alpha e_i y + \beta}{e_i^2}.\]

This problem occurs in HDR imaging, when estimating the unknown irradiance given noisy acquisitions with varying exposure times \cite{granados2010optimal,aguerrebere2014best}. Each $z_i$ is an unbiased estimator of $y$. Therefore, if the variances were known, we can minimize the MSE with the following weighted average, where the weights are the inverse of the variances:
\begin{equation}
\hat y = \frac{\sum_i w_i z_i}{\sum_i w_i},\quad
w_i = \frac{e_i^2}{\alpha e_i y + \beta}.
\label{eq:weighted_average}
\end{equation}
The problem is that the weights depend on the unknown $y$.
In \cite{granados2010optimal} Granados et al. solve this problem
with an iterative weighted average:
\begin{align*}
w_i^0 &= \frac{e_i^2}{\alpha e_i z_i + \beta}.\\
w_i^k &= \frac{e_i^2}{\alpha e_i \hat y^k + \beta},\quad \hat y^{k+1} = \frac{\sum_i w_i^k z_i}{\sum_i w_i^k},\quad k = 1,2,...
\end{align*}
It can be shown that this converges to the maximum likelihood estimate.

In our case, we are going to simplify expression \eqref{eq:weighted_average} by assuming that $\alpha e_i y \gg \beta$, and therefore $w_i \approx \frac{e_i}{\alpha y}$. Under this assumption, we obtain
\begin{equation}
\hat y = \frac{\sum_i e_i z_i}{\sum_i e_i}.
\end{equation}
This assumption holds for brighter pixels and well exposed images \cite{aguerrebere2014best}.

\section{HDR-DSP architecture}

Our HDR-DSP architecture has 3 trainable modules: Motion estimator, Encoder, and Decoder. The Feature Shift-and-Pool block does not have any trainable parameters. Our motion estimator follows the work of~\cite{sajjadi2018frame}. Our encoder and our decoder are inspired from the SRResNet architecture~\cite{ledig2017photo}, and built from the residual blocks (see Table~\ref{tab:Resblock}). Convolutions of the encoder and decoder are performed using reflection padding. In total, our networks have 2853411 trainable parameters (Table~\ref{tab:Architecture}).

\begin{table}[h]
\caption{\texttt{ResBlock(N)}}
\begin{center}
\resizebox{0.75\columnwidth}{!}{%
\begin{tabular}{l c}
\toprule
Input & Tensor N channels \\
Layer 1 & \texttt{Conv2d(in=N, out=N, k=3, s=1, p=1)}  \\
 & \texttt{ReLU}  \\
Layer 2 & \texttt{Conv2d(in=N, out=N, k=3, s=1, p=1)} \\
Output & Layer 2 + Input \\
\bottomrule
\end{tabular}
}
\label{tab:Resblock}
\end{center}
\end{table}

\begin{table}[ht]
\caption{HDR-DSP Architecture}
\setlength{\tabcolsep}{1pt}
\begin{center}
\resizebox{\columnwidth}{!}{%
\begin{tabular}{lcr}
\toprule
\bf Modules & \bf Layers & \bf Nb parameters\\
\midrule
Motion Estimator & \texttt{FNet}~\cite{sajjadi2018frame} & 1744354\\
\midrule
Encoder & \texttt{Conv2d(in=1, out=64, k=3, s=1, p=1)} & 332992\\
& \texttt{ResBlock(64)} $\times 4$ & \\
& \texttt{ReLU} & \\
& \texttt{Conv2d(in=64, out=64, k=3, s=1, p=1)} & \\
\midrule
FS\&P &  & 0\\
\midrule
Decoder & \texttt{Conv2d(in=64, out=64, k=3, s=1, p=1)} & 776065\\
& \texttt{ResBlock(64)} $\times 10$ & \\
& \texttt{ReLU} & \\
& \texttt{Conv2d(in=64, out=1, k=3, s=1, p=1)} & \\
\midrule
&  &  Total: 2853411\\
\bottomrule
\end{tabular}
\label{tab:Architecture}
}
\end{center}
\end{table}

\section{Training details} \label{Sec:Training}
We train \DSP in two stages, first pretraining the motion estimator and then the end-to-end system.

\paragraph{Phase 1: Pre-train the Motion Estimator.} 
Training the motion estimator in the case of images obtained with different exposures is a challenging task. 
We first pretrain it on the simulated dataset to ensure that it produces accurate flows. We monitor the quality of the estimations by comparing with the ground truth flows, until reaching an averaged error of 0.05 pixel.

For training the motion estimator our first choice was to use the $L_1$ distance between the  reference image and the radiometrically corrected warped image. However, the quality of the estimated flows were not acceptable (with errors above 0.1 pixel). 
Indeed, since motion estimation relies on the photometric consistency between frames, it is very sensitive to the intensity fluctuations between frames (as it is the case for our normalized LR frames $I_i^{LR}$), which results in imprecise alignments. 

To prevent this issue we compute the warping loss on the details rather than on the images, which is common in traditional optical flow~\cite{sajjadi2018frame, Liu_2019_CVPR}. 
The loss is computed for each flow $F_{i\to r}$ %
estimated by the $\textbf{MotionEst}$ module
\begin{multline}
\textstyle      \ell_{me}(\{F_{i\to r}\}_{i=1}^m) =  \\
\sum_i \| \textbf{Detail}(I_i^{LR}) - \textbf{Detail}\left(\textbf{Pullback} ( I_r^{LR}, F_{i\to r} ) \right) \|_1 \\+ \lambda_1 TV(F_{i\to r}),
\label{eq:meloss}
\end{multline}
where $\textbf{Pullback}$ computes a  bicubic warping of $I_r^{LR}$ according to a flow, $\textbf{Detail}$ applies a high-pass filter to the images, TV is the finite difference discretization classic Total Variation regularizer~\cite{rudin1992nonlinear}, and $\lambda_1=0.003$ is a hyperparameter controlling the regularization strength.

We set the batch size to 32 and use Adam~\cite{kingma2014adam} with the default Pytorch parameters and a initialized learning rate of $10^{-4}$ to optimize the loss. The pre-training converges after 50k iterations and takes about 3 hours on one NVIDIA V100 GPU.

\paragraph{Phase 2: Train the whole system end-to-end.} 
We then use the pretrained motion estimator and train the entire system end-to-end using the total loss:
\begin{equation}
    \text{loss} = \ell_{self} + \lambda_2 \ell_{me}.
    \label{eq:total-loss-self}
\end{equation}
We set $\lambda_2 = 3$ in our experiments. 
Furthermore, to avoid boundary issues, the loss does not consider values at a distance below 2 pixels from the border of the frames.

We train our model on LR crops of size $64 \times 64$ pixels and validate on LR images of size $256 \times 256$ pixels. During training, our network is fed with a random number of LR input images (from $4$ to $14$) in each sequence. We set the batch size to $16$ and optimize the loss using the Adam optimizer with default parameters.
The learning rates are initialized to $10^{-4}$, then scaled by 0.3 each 400 epochs. The training takes 20h (1200 epochs) on one NVIDIA V100 GPU.

\section{Trainable feature pooling alternative}

The feature pooling block FSP described in Section 4.1 of the main article does not have any trainable parameters. In this section we investigate the use of a trainable layer, named \AxelNet, for performing this task. %

We considered a simple trainable network \AxelNet that performs feature pooling (Table~\ref{tab:AxelNet}) instead of statistical feature poolings (Avg-Max-Std) as presented in the paper. To this aim, \AxelNet takes as input the concatenation of $N$ features $J_i^{HR}$ and $N$ weights $W_i^{HR}$ (computed by the SPMC~\cite{tao2017detail} module from $N$ LR images) and produces the fused HR features. Then, the Decoder network reconstructs the HR detail image from the fused features.  

A drawback of \AxelNet is that it can  only be applied on a fixed number of frames.  Table~\ref{tab:variableNumber} compares the performance of the \AxelNet (which replaces the Avg-Max-Std feature pooling) trained on 4 and 14 frames with our original method. We can see that in the case of small number of frames, \AxelNet attains a performance comparable to our \DSP method using the Avg-Max-Std feature pooling. However, in the case of 14 frames, there is a big gap of 0.3 dB between our method and \AxelNet. It seems that it is more difficult for \AxelNet to capture the necessary statistics from many features.

\begin{table*}[h]
\caption{\texttt{PoolNet($N$)} architecture for trainable feature pooling.}
\small
\begin{center}
\begin{tabular}{l c}
\toprule
Input & $N$ Features (64 channels) + $N$ Weights (1 channel) \\
 & \texttt{Conv2d(in=N(64+1), out=256, k=1, s=1, p=0)}  \\
  & \texttt{ReLU}  \\
 & \texttt{Conv2d(in=256, out=128, k=1, s=1, p=0)}  \\
  & \texttt{ReLU}  \\

 & \texttt{Conv2d(in=128, out=64, k=1, s=1, p=0)}  \\
  & \texttt{ReLU}  \\

Output & fused HR feature \\

\bottomrule
\end{tabular}
\label{tab:AxelNet}
\end{center}
\end{table*}

\begin{table*}%
\begin{center}
\small
\caption{Trainable feature pooling evaluation. Since the trainable feature pooling networks \AxelNet only accepts a fixed number of frames (in this case 4 and 14) we compare it with \DSP also trained with fixed number of frames. }
\setlength{\tabcolsep}{6pt}
\begin{tabular}{l ccccc}
\toprule
Methods & \DSP &  \DSP 4 & \DSP 14 & \AxelNet 4 & \AxelNet 14\\
\midrule
PSNR(dB) 4 frames & \textbf{52.81}  & 52.69 & 51.31 & 52.76 & N/A\\ 
PSNR(dB) 14 frames & \textbf{55.85}  & 54.26 & 55.53 & N/A & 55.55\\ 
PSNR(dB) variable $n$ frames & \textbf{54.70} & 53.85 & 54.07 & N/A & N/A \\ 

\bottomrule
\end{tabular}
\label{tab:variableNumber}
\end{center}
\end{table*}

\section{Alternative exposure weighting strategies }

As discussed in the main paper, the LRs with longer exposure time should contribute more to the reconstruction because of their high signal-to-noise ratio. In our proposed method, we use the un-normalized LR images as additional input to the Encoder so as that the Encoder perceives the noise level in each LR image. Subsequently, the Encoder can decide which features are more important.  

We also evaluated an alternative strategy to weight the features (WF) based on the exposure times. This simply consists in weighting the features $J_i^{LR}$ by the corresponding exposure time in the SPMC module. Actually, this was inspired from the ME S\&A method. 

This strategy leads to slightly worse yet adequate feature encodings (-0.08dB) as shown in Table~\ref{tab:exposureWeighting}. Moreover, using both feature weighting and LRs encoding (third column) leads to the same performance as only using LRs encoding. This implies that the Encoder already encodes the necessary information about the signal-dependent noise on the features.

\begin{table}[h]
\begin{center}
\footnotesize

\caption{Handling of the signal-dependent noise}
\label{tab:exposureWeighting}
\resizebox{\columnwidth}{!}{%
\begin{tabular}{l ccc}
\toprule
Methods & \DSP & DSP (+WF - LR) & DSP (+WF)\\
\midrule
PSNR(dB) ME & \textbf{54.70} & 54.62 & 54.70 \\ 
\bottomrule
\end{tabular}
}

\end{center}
\end{table}

\section{Adaptation of existing methods to multi-exposure sequences}

We detail here the adaptations to the algorithms we used in the comparisons.

\paragraph{ME S\&A.} \emph{Multi-exposure Shift-and-add} is a weighted version of the classic shift-and-add method~\cite{fruchter2002drizzle,merino2007super,grycewicz2008focal,jia2012method} designed for multi-exposure sequences. Usually, S\&A produces the HR image by registering the LR images onto the HR grid using the corresponding optical flows. After the registration step, the intensities of the LR images are splatted to the neighborhood integer-coordinate pixels using some kernel interpolation. Finally, pixel-wised aggregation is done to obtain the HR output image. Therefore a naive method consists of using the classic S\&A method on the normalized LR images. However this ignores the different signal-to-noise ratios in the normalized images and fails to greatly reduce the noise. Using the same arguments as in the Sec.~\ref{Sec:granados}, we propose the weighted S\&A for multi-exposure sequence
\begin{equation}
    \widehat I^{HR} = \dfrac{\sum_{i = 1}^m \textbf{Register}(\bar I_i^{LR})}{\sum_{i = 1}^m e_i}
\end{equation}
where \textbf{Register} maps and splats the un-normalized images $\bar I_i^{LR}$ onto the HR grid.

\paragraph{Base-detail ACT (BD ACT).}  
ACT~\cite{anger2020fast} is a traditional multi-image super-resolution method developed for Planet SkySat single-exposure sequences. It formulates the reconstruction as an inverse problem and solves it by an iterative optimization method.
BD ACT extends ACT to support multi-exposure images by adopting the same base-detail strategy as proposed in \DSP: the details of the images are fused by ACT, and the base is reconstructed by the upsampled average of the bases of the input images.

\paragraph{HighRes-net (HR-net) and RAMS.} 
HighRes-net~\cite{deudon2002highres} and RAMS~\cite{salvetti2020multi} are two super-resolution methods for multi-temporal PROBA-V satellite images. However in the PROBA-V dataset, the identity of the LR reference image is unavailable. This hinders the true potential of the methods trained on this dataset. As a result we use the reference-aware super-resolution~\cite{nguyen2021proba} of HighRes-net and RAMS. In HighRes-net, the reference image is used as a shared representation for all LR images. Each LR image is embedded jointly with this reference before being recursively fused. In RAMS, each LR image is aligned to the reference image before being input to the residual attention block. The registration step of RAMS is done with inverse compositional algorithm~\cite{briand2018improvements}, which is robust to noise and brightness change. 
As HighRes-net and RAMS are supervised methods, we also use a radiometric correction on the output before computing the loss~\cite{deudon2002highres}.

\paragraph{DSA.} \emph{Deep shift-and-add} \cite{Nguyen_2021_CVPR} 
DSA is a self-supervised method for super-resolution of push-frame single-exposure satellite images. We adapt DSA to multi-exposure case by using the normalized LR images as input. We also use the loss on the details to train the motion estimator in DSA.

\section{Execution time}
Table~\ref{tab:execution-time} reports the execution time of the methods studied on the synthetic multi-exposure dataset.
Due to its convolutional architecture, HighRes-net is the fastest. \DSP is slightly more costly than DSA since it performs feature pooling instead of a simple average and requires fusing the bases together. ME S\&A and BD-ACT are both executed on CPU, the later being quite costly due to the linear spline system inversion.

\begin{table*}%
\begin{center}
\small
\caption{Execution time (s) on 200 sequences of size $15 \times 256 \times 256$ pixels.}\label{tab:execution-time}
\begin{tabular}{l cccccc}
\toprule
Methods & RAMS & ME S\&A & HR-net & BD-ACT & DSA &HDR-DSP\\
\midrule
Time (s)  & 93 & 276 & 22 & 555 & 82 & 97 \\ 

\bottomrule
\end{tabular}
\end{center}
\end{table*}

\section{Additional comparisons using real SkySat sequences}

\begin{figure*}

\centering

\def\s{0.1065}
    \def\ss{0.195}

    \begin{subfigure}{\s\linewidth}
    \caption*{$LR_1$}
    \includegraphics[width=\linewidth]{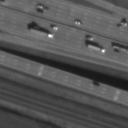}
    \end{subfigure}
    \begin{subfigure}{\s\linewidth}
    \caption*{$LR_2$}
    \includegraphics[width=\linewidth]{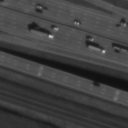}
    \end{subfigure}
    \begin{subfigure}{\s\linewidth}
    \caption*{$LR_3$}
    \includegraphics[width=\linewidth]{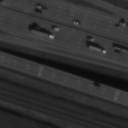}
    \end{subfigure}
    \begin{subfigure}{\s\linewidth}
    \caption*{$LR_4$}
    \includegraphics[width=\linewidth]{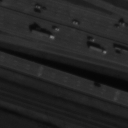}
    \end{subfigure}
    \begin{subfigure}{\s\linewidth}
    \caption*{$LR_5$}
    \includegraphics[width=\linewidth]{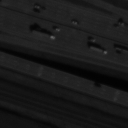}
    \end{subfigure}
    \begin{subfigure}{\s\linewidth}
    \caption*{$LR_6$}
    \includegraphics[width=\linewidth]{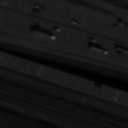}
    \end{subfigure}
    \begin{subfigure}{\s\linewidth}
    \caption*{$LR_7$}
    \includegraphics[width=\linewidth]{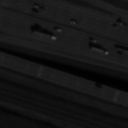}
    \end{subfigure}
    \begin{subfigure}{\s\linewidth}
    \caption*{$LR_8$}
    \includegraphics[width=\linewidth]{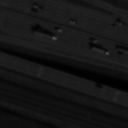}
    \end{subfigure}
    \begin{subfigure}{\s\linewidth}
    \caption*{$LR_9$}
    \includegraphics[width=\linewidth]{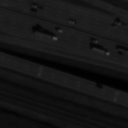}
    \end{subfigure}

\vspace{0.3cm}

\begin{tikzpicture}[every node/.style={inner sep=0,outer sep=0},spy using outlines={white,magnification=3,size=0.19\textwidth}]

  \newlength{\nextfigheight}Super-resolution from a real multi-exposure sequenc
  \newlength{\figwidth}
  \newlength{\figsep}
  \setlength{\nextfigheight}{0cm}
  \setlength{\figwidth}{0.19\textwidth}
  \setlength{\figsep}{0.2\textwidth}

\node[anchor=west, inner sep=0] (me) at (0*\figsep,\nextfigheight){\includegraphics[width=\figwidth]{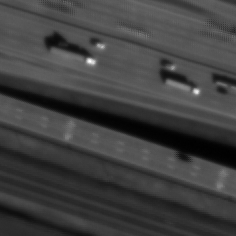}};
\node[anchor=west, inner sep=0] (planet) at (1*\figsep,\nextfigheight){\includegraphics[width=\figwidth]{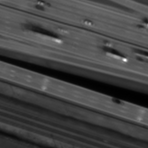}};
\node[anchor=west, inner sep=0] (act) at (2*\figsep,\nextfigheight){\includegraphics[width=\figwidth]{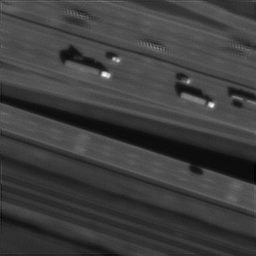}};
\node[anchor=west, inner sep=0] (dsa) at (3*\figsep,\nextfigheight){\includegraphics[width=\figwidth]{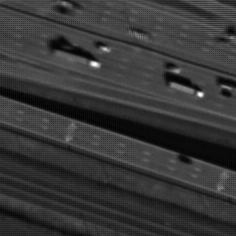}};
\node[anchor=west, inner sep=0] (dsp) at (4*\figsep,\nextfigheight){\includegraphics[width=\figwidth]{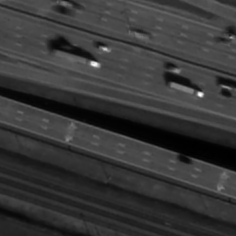}};

 \coordinate (spypoint_1) at (1.7cm          , 1.0cm);
 \coordinate (spypoint_2) at (1.7cm+  \figsep, 1.0cm);
 \coordinate (spypoint_3) at (1.7cm+2*\figsep, 1.0cm);
 \coordinate (spypoint_4) at (1.7cm+3*\figsep, 1.0cm);
 \coordinate (spypoint_5) at (1.7cm+4*\figsep, 1.0cm);

\addtolength{\nextfigheight}{-\figsep}

\spy on (spypoint_1) in node [name=me_spy] at (me.south) [anchor=north];
\spy on (spypoint_2) in node [name=planet_spy] at (planet.south) [anchor=north];
\spy on (spypoint_3) in node [name=act_spy] at (act.south) [anchor=north];
\spy on (spypoint_4) in node [name=dsa_spy] at (dsa.south) [anchor=north];
\spy on (spypoint_5) in node [name=dsp_spy] at (dsp.south) [anchor=north];

\node [anchor=south] at (me.north) {\scriptsize \begin{tabular}{r} ME S\&A\end{tabular}};
\node [anchor=south] at (planet.north) {\scriptsize \begin{tabular}{r} Planet L1B~\cite{murthy2014SkySat}\end{tabular}};
\node [anchor=south] at (act.north) {\scriptsize \begin{tabular}{r} BD-ACT~\cite{anger2020fast}\end{tabular}};
\node [anchor=south] at (dsa.north) {\scriptsize \begin{tabular}{r} DSA~\cite{Nguyen_2021_CVPR}\end{tabular}};
\node [anchor=south] at (dsp.north) {\scriptsize \begin{tabular}{r} Our \DSP\end{tabular}};

 \end{tikzpicture}
\caption{Super-resolution from a real multi-exposure sequence of 9 SkySat images. Top row: Original low resolution images with different exposures. Middle row: Reconstructions from five methods, including ours trained with self-supervision (right). Bottom row: Zoom on a detail of the results.}
\label{fig:highway}
\end{figure*}

Figure~\ref{fig:highway} presents results obtained on real multi-exposure SkySat images using 9 frames. This is a challenging sequence as it contains moving vehicles. Note how the road markings are better seen in the \DSP result. However, since \DSP does not account for moving objects (the motion estimator only predicts smooth motion within a range of 5 pixels) the cars are blurry.

\begin{figure*}

\centering
\def\s{0.138}
    \def\ss{0.195}

    \begin{subfigure}{\s\linewidth}
    \caption*{$LR_1$}
    \includegraphics[width=\linewidth]{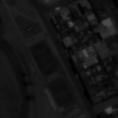}
    \end{subfigure}
    \begin{subfigure}{\s\linewidth}
    \caption*{$LR_2$}
    \includegraphics[width=\linewidth]{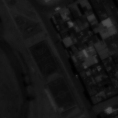}
    \end{subfigure}
    \begin{subfigure}{\s\linewidth}
    \caption*{$LR_3$}
    \includegraphics[width=\linewidth]{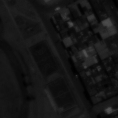}
    \end{subfigure}
    \begin{subfigure}{\s\linewidth}
    \caption*{$LR_4$}
    \includegraphics[width=\linewidth]{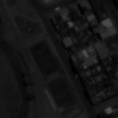}
    \end{subfigure}
    \begin{subfigure}{\s\linewidth}
    \caption*{$LR_5$}
    \includegraphics[width=\linewidth]{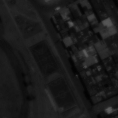}
    \end{subfigure}
    \begin{subfigure}{\s\linewidth}
    \caption*{$LR_6$}
    \includegraphics[width=\linewidth]{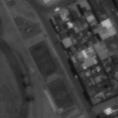}
    \end{subfigure}
    \begin{subfigure}{\s\linewidth}
    \caption*{$LR_7$}
    \includegraphics[width=\linewidth]{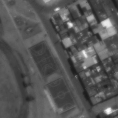}
    \end{subfigure}

\vspace{0.3cm}

\begin{tikzpicture}[every node/.style={inner sep=0,outer sep=0},spy using outlines={white,magnification=4,size=0.19\textwidth}]

  \setlength{\nextfigheight}{0cm}
  \setlength{\figwidth}{0.19\textwidth}
  \setlength{\figsep}{0.2\textwidth}

\node[anchor=west, inner sep=0] (me) at (0*\figsep,\nextfigheight){\includegraphics[width=\figwidth]{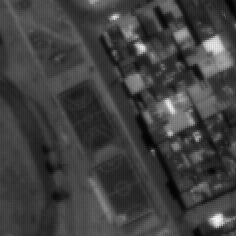}};
\node[anchor=west, inner sep=0] (planet) at (1*\figsep,\nextfigheight){\includegraphics[width=\figwidth]{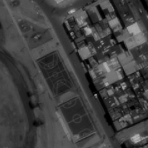}};
\node[anchor=west, inner sep=0] (act) at (2*\figsep,\nextfigheight){\includegraphics[width=\figwidth]{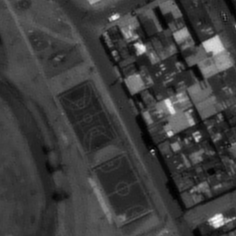}};
\node[anchor=west, inner sep=0] (dsa) at (3*\figsep,\nextfigheight){\includegraphics[width=\figwidth]{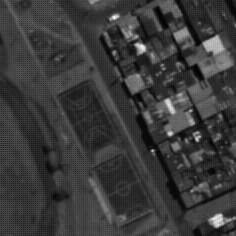}};
\node[anchor=west, inner sep=0] (dsp) at (4*\figsep,\nextfigheight){\includegraphics[width=\figwidth]{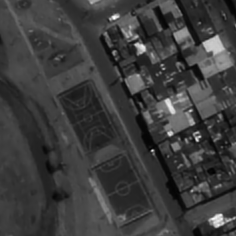}};

 \coordinate (spypoint_1) at (2.2cm, 0.2cm);
 \coordinate (spypoint_2) at (2.2cm+1*\figsep, 0.2cm);
 \coordinate (spypoint_3) at (2.2cm+2*\figsep, 0.2cm);
 \coordinate (spypoint_4) at (2.2cm+3*\figsep, 0.2cm);
 \coordinate (spypoint_5) at (2.2cm+4*\figsep, 0.2cm);

\addtolength{\nextfigheight}{-\figsep}

\spy on (spypoint_1) in node [name=me_spy] at (me.south) [anchor=north];
\spy on (spypoint_2) in node [name=planet_spy] at (planet.south) [anchor=north];
\spy on (spypoint_3) in node [name=act_spy] at (act.south) [anchor=north];
\spy on (spypoint_4) in node [name=dsa_spy] at (dsa.south) [anchor=north];
\spy on (spypoint_5) in node [name=dsp_spy] at (dsp.south) [anchor=north];

\node [anchor=south] at (me.north) {\scriptsize \begin{tabular}{r} ME S\&A\end{tabular}};
\node [anchor=south] at (planet.north) {\scriptsize \begin{tabular}{r} Planet L1B~\cite{murthy2014SkySat}\end{tabular}};
\node [anchor=south] at (act.north) {\scriptsize \begin{tabular}{r} BD-ACT~\cite{anger2020fast}\end{tabular}};
\node [anchor=south] at (dsa.north) {\scriptsize \begin{tabular}{r} DSA~\cite{Nguyen_2021_CVPR}\end{tabular}};
\node [anchor=south] at (dsp.north) {\scriptsize \begin{tabular}{r} Our \DSP\end{tabular}};

 \end{tikzpicture}
\caption{Super-resolution from a real multi-exposure sequence of 7 SkySat images. Top row: Original low resolution images with different exposures. Middle row: Reconstructions from five methods, including ours trained with self-supervision (right). Bottom row: Zoom on a detail of the results.}
\label{fig:basketball}
\end{figure*}

Figure~\ref{fig:basketball} shows another example of reconstruction on a real sequence of 7 SkySat images. Even though there are only 7 images in this sequence and most of them are very noisy, \DSP is able to produce a clean image. The fine details are well restored.

\section{Exposure error analysis}

\begin{figure}
    \centering
    \includegraphics[width=0.49\textwidth]{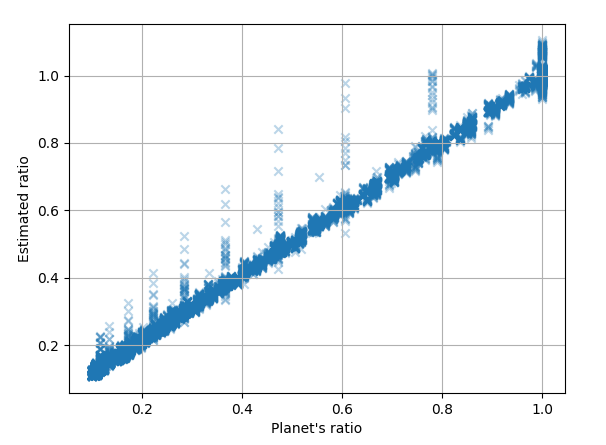}
    \caption{Normalized estimated exposure ratio with respect to provided exposure time.}
    \label{fig:plot-exposure}
\end{figure}

We observed a discrepancy between the reported exposure time by Planet and correct normalization ratios. This can be explained by measurement imprecision since the quantities are in sub-milliseconds range, or by local illumination effects such as vignetting.
To estimate the correct exposure ratio for a given pair of images, we registered the images using phase correlation, masked saturated pixels and computed the spatial median of the ratio between the two frames. We then validated visually that such exposure ratios were more precise that the reported exposure time (less flicker was observed).

Figure~\ref{fig:plot-exposure} shows the relation between the reported ratio, and the estimated one. We find that errors are usually in the order of a few percent, but also observe larger errors. The nominal exposure times range from 0.4ms to 4.5ms.
Note that the absolute error in exposure the time measurement is probably constant regardless of the exposure time. However, when computing the ratio of two exposures with errors this might result in a large divergence of the ratio, especially if the  exposure in the denominator is a short one.

Note that for the proposed super-resolution method, we used the imprecise, reported exposure times and not the estimated one, as the estimation method itself can fail.

\bibliographystyle{ieee_fullname}
\bibliography{egbib}